# Designing a Hair-Clip Inspired Bistable Mechanism for Soft Fish Robots*

Zechen Xiong[1], Hod Lipson[2]

*Abstract*—The Hair clip mechanism (HCM) is an in-plane prestressed bistable mechanism proposed in our previous research [1]–[5] to enhance the functionality of soft robotics. HCMs have several advantages, such as high rigidity, high mobility, good repeatability, and design- and fabrication-simplicity, compared to existing soft and compliant robotics. Using our experience with fish robots, this work delves into designing a novel HCM robotic propulsion system made from PETG plastic, carbon fiber-reinforced plastic (CFRP), and steel. Detailed derivation and verification of the HCM theory are given, and the influence of key parameters like dimensions, material types, and servo motor specifications are summarized. The designing algorithm offers insight into HCM robotics. It enables us to search for suitable components, operate robots at a desired frequency, and achieve high-frequency and high-speed undulatory swimming for fish robots.

## I. INTRODUCTION

Soft and compliant robotics is an emerging field that focuses on designing and constructing robots with soft or deformable materials. The inherent softness of the material used enables these robots to mimic natural movements and interact with their environment in a novel way [6], [7]. Soft robotic fish, in particular, have garnered significant attention due to their potential in non-intrusive underwater exploration and environmental monitoring [8]. Unlike traditional propeller-driven underwater vehicles, soft fish robots are usually driven by fluidic or electroactive polymer actuators and locomote through traveling sinusoidal waves that undulate their bodies. For example, Marchese et al. [9], Katzschmann et al.[10], Marchese et al. [11], and Katzschmann et al. [8] present the design, fabrication, control, and oceanic testing of a type of soft robotic fish that is self-contained and capable of rapid locomotion. Their fish robot swims in three dimensions and can monitor aquatic life and environments. Li et al. [12] and Li et al. [13] design and fabricated a ray fish-inspired soft swimmer actuated by dielectric-elastomer that is able to locomote at high speed and sustain high water pressure in the Mariana trench. Zhang et al. [14] study global vision-based formation control of soft robotic fish swarm.

While soft robotic fish show great promise, several challenges remain. First of all, the highest speed achieved by this type of fish robot is only about 0.5 [8] ~ 0.7 BL/s [12]. Others may have demonstrated faster speed but are not adequate in a tetherless situation [12], [15], [16]. Second, the performance of such robots is significantly influenced by their empirical design and manual fabrication. In recent years, it has been noted that energy-storing-and-releasing mechanisms can significantly increase the moving speed and strength of robots and organisms, addressing the low-velocity problem of soft robotics. For instance, Hawkes et al. [17] designed and fabricated a jumping robot that stores and releases multiplied work to achieve a height of 30 meters; Pal et al. [18] present prestressed pneumatic soft actuators that recover at a short timescale of ~ 50 ms through energy storing and releasing. Smith et al. [19] study the beaks of hummingbirds that close in tens of milliseconds through an appropriate sequence of muscle actions. Forterre et al. [20] analyze the rapid closure of the Venus flytrap and show that the snap-buckling instability plays an important role. Meanwhile, the use of semi-rigid materials, 2D materials, and 3D-printing technologies profoundly changed the design and fabrication methodology of soft robotics [6], [21].

We propose to use prestressed bistable mechanisms to solve these problems. In our previous work [2]–[4], a type of in-plane prestressed bistable hair-clip-like mechanism is designed, fabricated, and analyzed, which we term hair clip mechanism (HCM, Figure 1). HCMs have a stiffness about 8-11 times higher than their unassembled precursor [3], [22] (we borrow the terms "precursor" and "postcursor" to distinguish between the "unassembled" and "assembled" HCM ribbons, respectively), which enables their usage as the robotic skeleton and motion-transmission mechanism at the same time. The bistability of HCMs unlocks their function as force or power amplifiers that improve the performance of soft robots. For example, we demonstrated a tetherless robotic fish that is capable of swimming at a speed of 2.03 BL/s [2] and a galloping robotic crawler that has a speed of 1.56 BL/s [3]. Both are among the highest rates achieved among their kindred. Besides, most HCM parameters can be derived from an approximated mathematical model for HCMs with acceptable errors, which provides us with a systematic design algorithm for HCM soft robots.

The idea of using snap-through buckling mechanisms as robotic propulsion also appeared in others' work [23]–[26]. However, the structures used only show their force-amplifying function but don't have high efficiency or

*Research supported by the Fu Foundation School of Engineering and Applied Science, Columbia University and U.S. National Science Foundation (NSF) AI Institute for Dynamical Systems grant 2112085.

[1]Zechen Xiong is with the Dept. of Earth and Environment Engineering at Columbia University, New York, NY 10027 USA (phone: 9173023864, e-mail: zechen.xiong@columbia.edu).

[2]Hod Lipson is with the Dept. of Mechanical Engineering at Columbia University, New York, NY 10027 USA (e-mail: ys3399@columbia.edu and hod.lipson@columbia.edu).

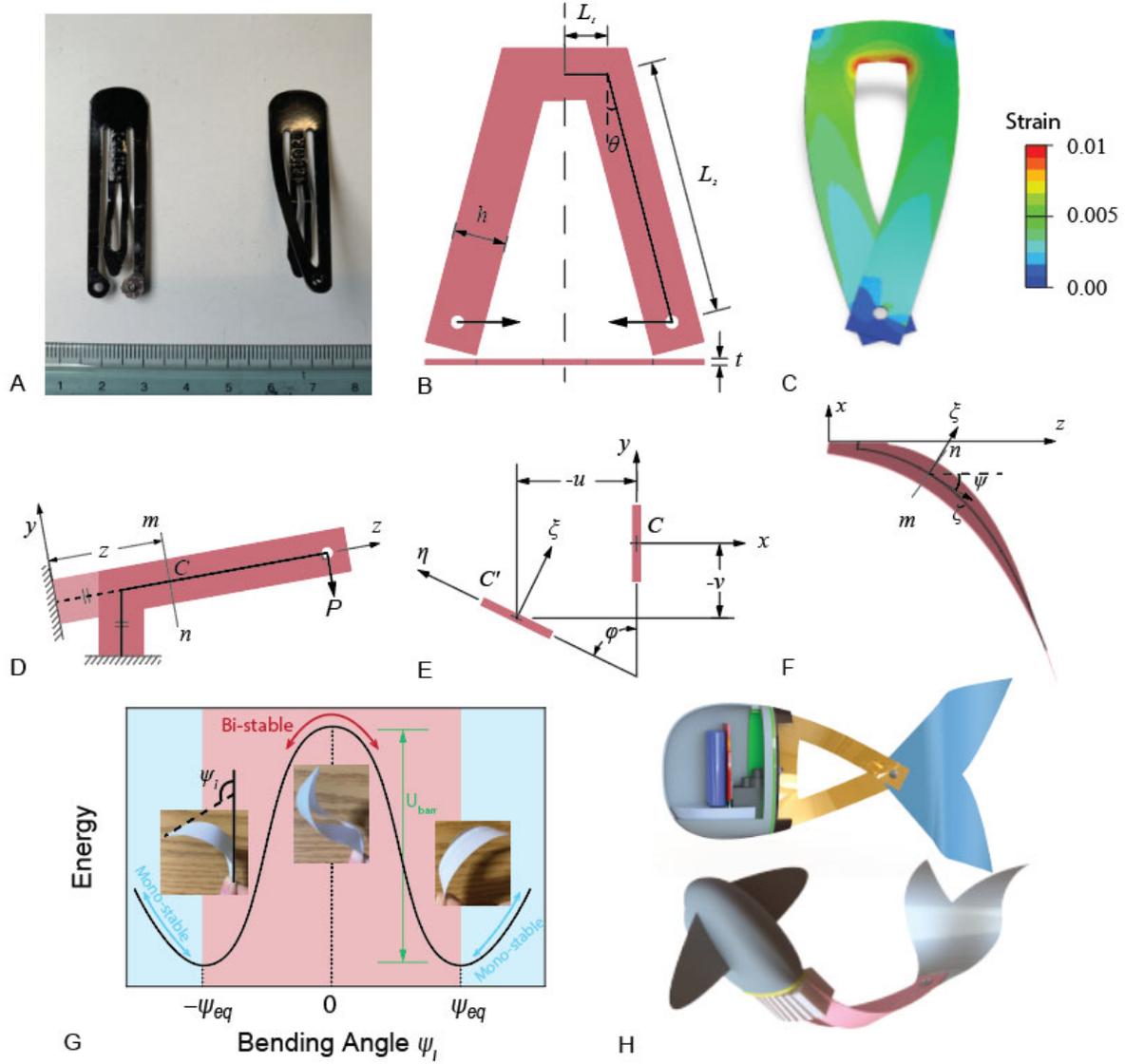

Figure 1 The working principle, mathematical model, and soft robotic applications of the hair-clip mechanism (HCM). (A) The configuration of a steel hair clip before and after assembly. (B) and (C) The assembly of a typical HCM. (D) - (F) The mathematical modeling of the hair clip mechanism (HCM). The coordinate $z$ is defined as the equivalent length of the half ribbon. (G) The energy landscape of bistable HCM. (H) The fish robots with HCM as fish body and tail [2].

convenience, nor a detailed theory or designing algorithm is proposed for their use. Also, most of them are in tethered situations. The use of HCM will probably solve these problems.

## II. WORKING PRINCIPLES

### A. HCM theory

The construction, operation, and application of the hair-clip mechanism (HCM) are depicted in Figure 1. By connecting the two extremities of an angled plastic strip, a bistable mechanism resembling a hair clip emerges. For mathematical solution derivation, we define the coordinate systems and variables in Figure 1B and 1D-1G. The displacement components $u$, $v$, and $\varphi$ are gauged at centroid $C$ of a random section $mn$ along the $x$, $y$, and $z$ axes (fixed space coordinates). The signs of them adhere to axis directions and the right-hand rule. The $\xi$, $\eta$, and $\zeta$ axes (follower coordinates) are established through the altered centroid $C'$ of section $mn$, aligning with the main axes of the deformed configuration. Using the small deflection assumption, Euler beam theory, and considering the angled ribbon as a direct cantilever beam with a constant rectangular cross-section (Figure 1D), the assembly process is expressed by mathematical equations below [27]:

$$EI_\xi \frac{d^2v}{dz^2} + P(l-z) = 0 \quad (1)$$

$$EI_\eta \frac{d^2u}{dz^2} + P\varphi(l-z) = 0 \quad (2)$$

$$-C\frac{d\varphi}{dz} + P(l-z)\frac{du}{dz} - P(u_1 - u) = 0 \quad (3)$$

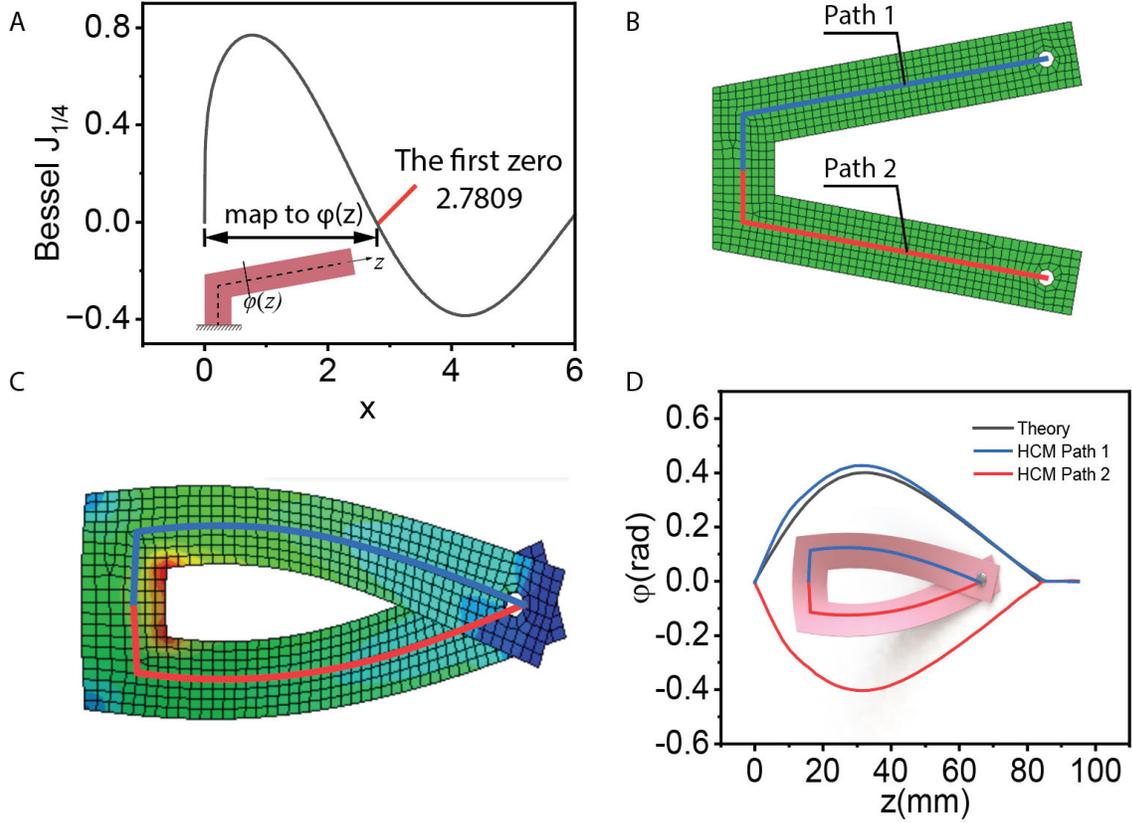

Figure 2 The validation of the variable $\varphi$ using Abaqus FE simulation. (A) The physical meaning of Bessel function $J_{1/4}$ and its map to rotation angle $\varphi$. (B) and (C) The FE method simulation of HCM with shape factor $\theta = 10°$, $\gamma_s = L_2 / L_1 = 6$, and other parameters $h / L_1 = 15$ mm / 12.5 mm and $t / L_1 = 0.381$ mm / 12.5 mm. (C) Comparison of the theoretical calculation and FE simulated value of $\varphi$.

Here, $EI_i = E \times (\text{width}_i) \times (\text{height}_i)^3/12$ represents the flexural rigidity along the $i$ axis (Figure 1D-1F), $l = L_1 + L_2$ signifies the strip's half-length, $C = GJ = hb^3G/3$ denotes the torsional rigidity of the thin section $mn$, $u_1$ is the horizontal deflection at the beam end $z = l$, $G = E/2/(1-v)$ is the shear modulus, and $v$ the Poisson ratio. Differentiating Eq. (3) concerning $z$ and incorporating in Eq. (2) gives

$$C\frac{d^2\varphi}{dz^2} + \frac{P^2(l-z)^2}{EI_\eta}\varphi = 0 \tag{4}$$

The general solution of Eq. (4) is:

$$\varphi = \sqrt{s}\left[A_1 J_{1/4}\left(\frac{\beta_1}{2}s^2\right) + A_2 J_{-1/4}\left(\frac{\beta_1}{2}s^2\right)\right] \tag{5}$$

where $s = l - z$, $J_{1/4}$ and $J_{-1/4}$ represent the Bessel functions of the first kind of order $1/4$ and $-1/4$, respectively, and

$$\beta_1 = \sqrt{\frac{P^2}{EI_\eta C}}. \tag{6}$$

Given the ribbon's symmetry, the small deflection premise, and the boundary condition:

$$\varphi|_{s=0} = 0, \tag{7}$$

we can infer that the integration constant $A_2$ equals 0, which reduces Eq. (5) to only one term:

$$\varphi(z) = \sqrt{l-z}\, A_1 J_{1/4}\left(\frac{1}{2}\sqrt{\frac{P_{cr}^2}{EI_\eta C}}(l-z)^2\right), \tag{8}$$

in which $A_1$ is a non-zero integration constant that can be determined from energy conservation. The value of $P_{cr}$ can be derived from the other boundary equation:

$$\varphi(0) = \sqrt{l}\, A_1 J_{1/4}\left(\frac{1}{2}\sqrt{\frac{P_{cr}^2}{EI_\eta C}}l^2\right) = 0 \tag{9}$$

Since $A_1$ is a non-zero constant, the $J_{1/4}$ term must be zero. At the same time, the physical meaning of the $J_{1/4}$ term is the configuration of the deformed ribbon (with a simple mapping), which should only have one zero point because of the lack of lateral support along the half-ribbon path. The profile and physical meaning of the Bessel function is shown in Figure 2A. Plugging the first zero into Eq. (9) yields

$$\frac{1}{2}\sqrt{\frac{P_{cr}^2}{EI_\eta C}}l^2 = 2.7809 \tag{10}$$

and thus,

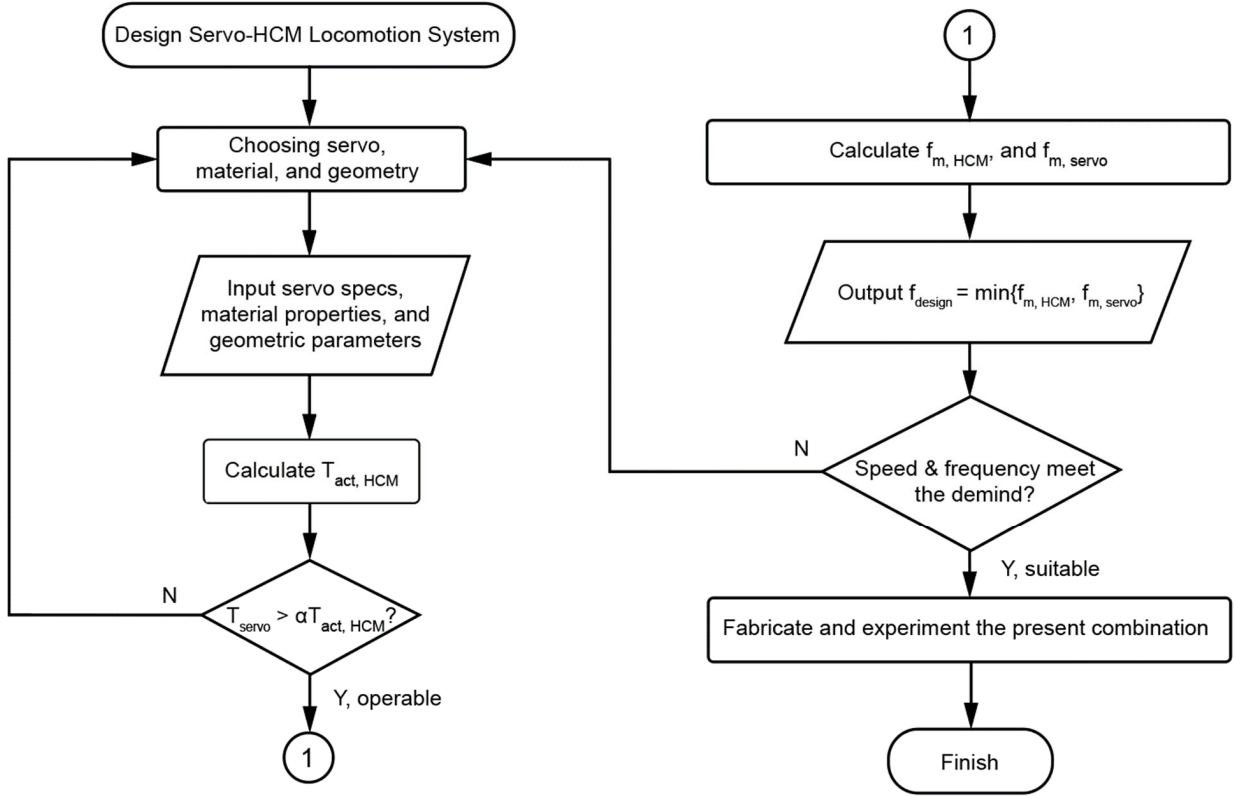

Figure 3 Designing algorithm of servo-HCM robotic systems based on HCM theory of Eq. (16)-(20). The design factor $\alpha$ is assumed to be 1.0 in our cases.

$$P_{cr} = \frac{5.5618}{l^2} \cdot \sqrt{EI_\eta C}. \tag{11}$$

Plugging Eq. (10) into Eq. (8) gives the analytical expression of

$$\varphi(z) = \sqrt{l-z}\, A_1 J_{1/4}\left(2.7809\left(\frac{l-z}{l}\right)^2\right). \tag{12}$$

To decide $A_1$'s value and subsequently $\varphi$, we incorporate an energy conservation equation, which states that the structure's total elastic strain energy must equate to the external load's work. With the membrane strain energy disregarded, we deduce that:

$$U = \frac{1}{2}\int_0^l \left[\frac{M_\eta^2}{EI_\eta} + GJ\left(\frac{d\varphi}{dz}\right)^2\right] dz$$

$$= \frac{1}{2}\int_0^l \left[P_{cr}^2 (l-z)^2 \sin^2\varphi / EI_\eta + GJ\left(\frac{d\varphi}{dz}\right)^2\right] dz \tag{13}$$

$$V = P_{cr} \cdot D = P_{cr} \cdot L_2\left(\sin^{-1}\frac{1}{\gamma_s} + \theta\right) \tag{14}$$

and

$$U = V \tag{15}$$

in which $U$ represents the elastic strain energy, $M_\eta$ is the beam's moment along the $\eta$ axis, $V$ signifies the external work by $P_{cr}$, and $D$ denotes the prestressing distance. The value of $A_1$ hinges on the value of $\gamma_s = L_2 / L_1$ and $\theta$ (Figure 1) and typically ranges between 0.09 and 0.10.

*B. Validation and application*

The verification of Eq. (12) is presented with the help of the Abaqus FE simulation (Figure 2B-2D). Finally, by integrating Eq. (12) into Eq. (2), we can determine section *mn*'s out-of-plane bending angle $\psi$ from

$$\psi(z) \approx \frac{du}{dz} = -\frac{P_{cr}}{EI_\eta}\int_0^z \varphi(l-z)\,dz \tag{16}$$

and the translational displacement

$$u(z) = \int_0^z \varphi(s)\,ds, \tag{17}$$

We validated Eq. (16) by comparing it with experiments and corresponding FE simulations in [2]. The energy barrier between the bi-states of the HCM can be approximated as [2]

$$U_{barr} = 3P_{cr} \cdot D \tag{18}$$

Assuming Hooke's law (linear elasticity), when the servo deforms HCM, the peak torque required to actuate the HCM can be calculated as

$$T_{act, HCM} = 2U_{barr} \cdot L_{horn} / 2u(L_1) \quad (19)$$

in which $L_1$ is the length denoted in Figure 1B, and $L_{horn}$ is the length of the horn that goes with the servo. Usually, the design operational frequency is the key parameter that determines the performance of the HCM robot. It can be calculated as

$$f_{design} = \min \begin{Bmatrix} f_{m, HCM} = 1/2t_*, \\ f_{m, servo} = speed / 4u(L_1) \end{Bmatrix}, \quad (20)$$

in which $t_*$ is the timescale of the HCM snapping that is estimated as [28]

$$t_* = \frac{(2l)^2}{t\sqrt{E/\rho_s}}. \quad (21)$$

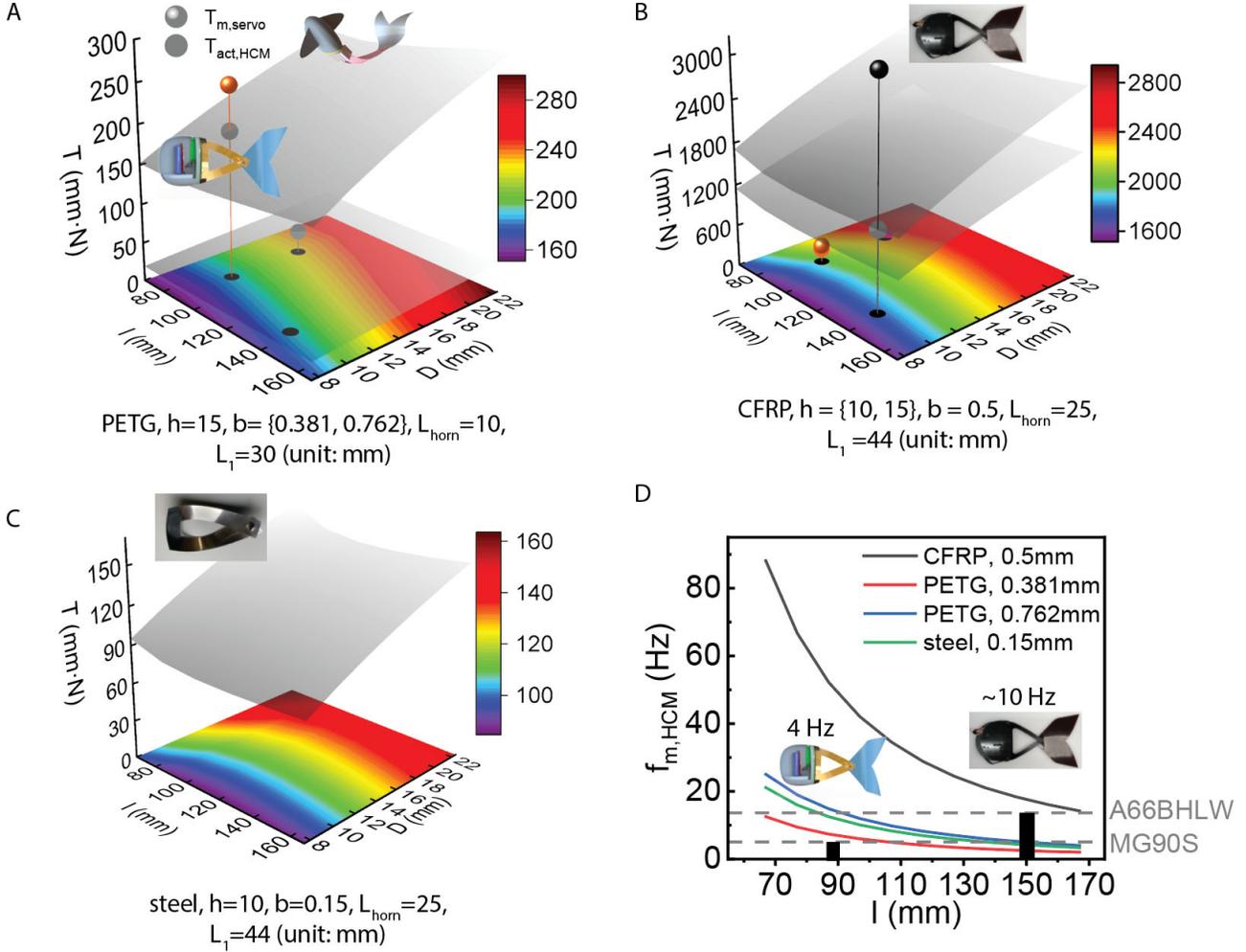

Figure 4. The influence of materials and dimensions to the actuation torque $T_{act, HCM}$ and the design frequency $f_{design}$. (A)-(C) The $T_{act, HCM}$ profile w.r.t half ribbon $l$ and prestressing distance $D$ of PETG plastic sheets, CFRP sheets, and steel sheets, respectively. The $T_{act, HCM}$ is found to be proportional to $D$, $b^3$, and $h$, but is less affected by $l$. Actual prototypes are pinpointed in the profiles for comparison and verification. (D) The design frequency $f_{design}$ calculation of different materials, geometry, and servos. Bothe the coral and black fish robots are limited by the speed of the servos.

## III. RESULTS

### A. Designing algorithm

According to the HCM theory, the algorithm for designing a servo-HCM locomotion system fish robots can be illustrated in Figure 3. Usually, the most important and time-consuming procedure is to look for correct combinations of servo, material, and shapes, according to our experience with the two fish robots in [2]. Typical materials include plastic, CFRP, and steel, whose related properties are given in Table I—their high tensile modulus and elastic limit enable them to withstand repeating loads and strain. The specifications of common servo motors are provided in Table II, in which $T_{servo}$ denotes the stall torque. A design factor of $\alpha > 1$ is suggested to ensure the successful servo-driven snap-through buckling of the HCM. In most situations, the speed of the fish robots is positively correlated with the undulating frequency. Thus, the iteration of HCM's geometry to configure the undulating frequency and strength offers a simple fix.

TABLE I. DIMENSIONS AND MECHANICAL PROPERTIES OF TYPICAL HCM MATERIALS

| Material | $t$ (mm) | $\rho$ (t/mm³) | $E$ (MPa) | $E/\rho$ (mJ/t) |
|---|---|---|---|---|
| PETG | 0.381, 0.762 | 1.25e-9 | 1.7e3 | 1.42e12 |
| CFRP | 0.5, 0.79 | 1.6e-9 | 64e3 | 40e12 |
| steel | 0.15 | 7.8e-9 | 200e3 | 25e12 |

TABLE II. SPECIFICATIONS OF COMMON SERVO MOTORS

|  | $T_{servo}$ (mm·N) | speed (rad/s) | weight (g) | $L_{horn}$ (mm) | $f_{m,servo}$ (Hz) |
|---|---|---|---|---|---|
| MG90S | 245 | 10.5 | 14 | 10 | 4.5 |
| B24CLM | 588 | 12.3 | 22 | 20 | 6.15 |
| A66BHLW | 3234 | 15.4 | 66 | 25 | 13.6 |
| A06CLS | 294 | 20.1 | 7 | 13 | 17.0 |
| DS3230MG | 3381 | 6.16 | 58 | 25 | 3.08 |
| SG92R | 245 | 10.5 | 9 | 10 | 4.5 |
| ZOSKAY | 3430 | 9.5 | 60 | 25 | 4.76 |

*B. Parametric analysis*

The designing logic and parametric influence of HCM fish robots are further discussed in Figure 4. The major elements that influence the peak torque ($T_{act, HCM}$) required to actuate the snapping-through of the HCM are material types and geometric dimensions half ribbon length $l$, the prestressing distance $D$, ribbon height $h$, ribbon thickness $t$, horn length $L_{horn}$, and HCM half width $L_1$. The required actuation torque of HCM is proportional to $D$, $h$, and $b^3$ yet is less relevant to the HCM ribbon length $2l$ since the increase of $l$ not only decreases energy barrier $U_{barr}$ but also decreases $u(L_1)$, the required actuation displacement of the HCM snap-through.

Three prototypes are juxtaposed in Figure 4A and 4B to compare the design theory with actual specifications. The pink and coral fish robots are made from PETG plastic HCMs, whose parameters are shown. The pink one has a geometry of $b = 0.381$ mm, $l = 87.5$ mm, and $D = 17.15$mm, which corresponds to a required torque of $T_{act, HCM} = 28.3$ mm·N; the coral fish robot has $b = 0.762$ mm, $l = 87$, $D = 11.75$mm, corresponding to $T_{act, HCM} = 188.7$ mm·N and thus a design factor $\alpha_{coral} = 245/188.7 = 1.30$ due to the use of an MG90S. The black fish robot prototype is made from CFRP HCM with $h = 10$ mm, $l = 137$ mm, and $D = 10$ mm, and thus a design factor $\alpha_{carbon} = 3234/1177 = 2.75$ with A66BHLW as the driving servo. The $T_{act,HCM}$ profile of steel HCM is plotted in Figure 4C.

The frequency analysis is illustrated in Figure 4D. It is shown that the limiting condition is usually the speed of the servos: the coral fish has a maximum HCM frequency of $f_{m, HCM} = 14.8$ Hz, while the MG90S servo operates at a maximum of $T_{m,servo} = 4.5$ Hz. The actual undulation of it is 0 ~ 4 Hz [2]. The CFRP fish robot has $f_{m, HCM} = 21.1$ Hz and $T_{m,servo} = 13.6$ Hz, and the actual operation is 0 ~ 10 Hz. To further increase the flapping frequency and swimming speed in the future, a DC motor-driven HCM system would be very beneficial.

## IV. CONCLUSION

The research presented delves into the innovative application of the Hair clip mechanism (HCM) to enhance the capabilities of soft robotics. The HCM, an in-plane prestressed bistable mechanism, offers a myriad of advantages, including high rigidity, mobility, repeatability, and ease of design and fabrication. This work, building upon our prior research, explores the design of an HCM robotic propulsion system using materials such as PETG plastic, CFRP, and steel. Through comprehensive analysis, the influence of various parameters like dimensions, material types, and servo motor specifications on the HCM's performance is elucidated.

Our findings underscore the potential of HCMs in addressing the challenges faced by soft robotic fish, particularly in achieving higher speeds and more efficient locomotion. The systematic design algorithm provided for HCM robots, backed by mathematical models, paves the way for more predictable and efficient robot designs. Future endeavors could explore the integration of DC motor-driven HCM systems to amplify further the flapping frequency and swimming speed of robotic fish.